# Nyström Kernel Mean Embeddings

Antoine Chatalic[1]   Nicolas Schreuder[1]   Alessandro Rudi[2]   Lorenzo Rosasco[1 3]

## Abstract

Kernel mean embeddings are a powerful tool to represent probability distributions over arbitrary spaces as single points in a Hilbert space. Yet, the cost of computing and storing such embeddings prohibits their direct use in large-scale settings. We propose an efficient approximation procedure based on the Nyström method, which exploits a small random subset of the dataset. Our main result is an upper bound on the approximation error of this procedure. It yields sufficient conditions on the subsample size to obtain the standard $n^{-1/2}$ rate while reducing computational costs. We discuss applications of this result for the approximation of the maximum mean discrepancy and quadrature rules, and illustrate our theoretical findings with numerical experiments.

## 1. Introduction

Owing to the increasing complexity of modern datasets, designing compact and meaningful representations of data collections and probability distributions is a key problem in machine learning. Kernel methods, which have proven in the last decades to be a convenient and powerful tool to design feature spaces capturing complex relations between data points, can be used towards this goal. Initially introduced by Smola et al. (2007), kernel mean embeddings (KME) indeed allow to represent a probability distribution via a single mean element in such feature spaces (Muandet et al. 2017). They have found applications in various areas such as anomaly detection (Zou et al. 2014), approximate Bayesian computation (Park et al. 2016), domain adaptation (Zhang et al. 2013), imitation learning

(Kim et al. 2018), nonparametric inference in graphical models (Song et al. 2013), functional data analysis (Hayati et al. 2020), discriminative learning for probability measures (Muandet et al. 2012) and differential privacy (Balog et al. 2018; Chatalic et al. 2021).

In machine learning applications, one is typically interested in computing the kernel mean embedding of the data distribution, which is most often unknown and approximated using empirical data. Yet, in its basic form, the empirical estimator still requires to store and manipulate the whole dataset, which can be impractical or even, in the worst case, impossible.

Finite-dimensional approximations of the feature map can be used to reduce the memory requirement of this approach (Liu et al. 2021; Rahimi et al. 2008). However, these methods are either generic and thus fail to capture the complexity of the data distribution with a low-dimensional representation, or adaptive, in which case costly data-dependent quantities need to be computed and/or stored (e.g., leverage scores (Shahrampour et al. 2019)). Furthermore, the resulting embeddings belong to a different space than the kernel mean embeddings and thus these approaches cannot be compared directly.

In this paper, we tackle the problem of efficiently approximating the kernel mean embedding of an unknown probability distribution for which samples are available. We introduce an estimator based on Nyström approximation, which can be computed efficiently using a random subset of the data, and provide an upper bound on its approximation error. Under mild assumptions on the kernel and data distribution, we show that the size of this random subset is much smaller than the number of available samples and yields an optimal error rate. We characterize this fact using the spectral properties of the covariance operator associated to the considered kernel. For instance, assuming an exponential decay of this spectrum, we show that a subsample size of the order of $\sqrt{n}\log(\sqrt{n})$ is sufficient to obtain an error of order $O(1/\sqrt{n})$, that is the same as that of the empirical estimator, and thus demonstrating that it is possible to preserve statistical performance while drastically re-ducing computational costs. We then show that our main result can be used to derive bounds on the approximation of the maximum mean discrepancy, which is the standard metric between probability distributions in the context of kernel methods. Finally we present numerical experiments supporting our theory both on synthetic and real datasets.

The paper is organized as follows. We begin by formalizing the problem of mean embeddings estimation and presenting related works in Section 2. Then, in Section 3, we introduce our Nyström estimator and detail its computational aspects. Section 4 contains our main result – a bound on the approximation error of our estimator. We discuss applications of this result in Section 5. Finally, we present the results of our numerical experiments in Section 6 and provide some perspectives in Section 7.

## 2. Mean embeddings estimation

Let $\rho$ be a probability measure with support in some space $\mathcal{X}$. We consider the problem of approximating its kernel mean embedding

$$\mu(\rho) := \int_{\mathcal{X}} \phi(x) \, \mathrm{d}\rho(x) \tag{1}$$

where $\phi : \mathcal{X} \to \mathcal{H}$ is a feature map taking values in a Hilbert feature space $\mathcal{H}$ with inner product $\langle \cdot, \cdot \rangle_{\mathcal{H}}$ and norm $\|\cdot\|$. By abuse of notation, we denote by $\mu = \mu(\rho)$ the embedding of the data distribution. We assume that $\phi$ is bounded so that the integral is well defined, and provide further technical details in Section 4. The choice of the feature map $\phi$ implicitly defines a positive definite kernel function $\kappa : \mathcal{X} \times \mathcal{X} \to \mathbb{R}$ via the equation $\kappa(x, y) = \langle \phi(x), \phi(y) \rangle_{\mathcal{H}}$. Strictly speaking, a kernel mean embedding corresponds to the case where $\phi$ is the canonical feature map of the kernel, i.e. $\phi(x) = \kappa(x, \cdot)$, however (1) can be defined for any integrable feature map and our results hold in this general setting.

**Maximum mean discrepancy** Mean embeddings naturally induce a semi-metric on the space of probability distributions $\mathcal{P}(\mathcal{X})$ known as the maximum mean discrepancy (Smola et al. 2007) and defined as

$$\mathrm{MMD}(\rho_1, \rho_2) := \|\mu(\rho_1) - \mu(\rho_2)\|,$$

for any two distribution probabilities $\rho_1$ and $\rho_2$. It satisfies all the properties of a metric except, in general, the definiteness, depending on whether the mean embedding $\rho \mapsto \mu(\rho)$ is injective or not (we refer the interested reader to Sriperumbudur et al. (2010) for more details). Such metrics have found applications in many contexts such as, to cite a few, two-sample testing (Borgwardt et al. 2006; Gretton et al. 2012), neural networks optimization (Borgwardt et al. 2006), generative models (Li et al. 2017; Sutherland et al. 2017). Given their wide applicability, maximum mean discrepancies are naturally one of the main motivations for better approximating mean embeddings. An interesting property of the MMD is that it is an integral probability metric (Müller 1997), a class of metrics which uses test functions to compare distributions. More precisely, we have

$$\mathrm{MMD}(\rho_1, \rho_2) = \sup_{f \in \mathcal{H} : \|f\| \leq 1} |\mathbb{E}_{X_1 \sim \rho_1} f(X_1) - \mathbb{E}_{X_2 \sim \rho_2} f(X_2)|$$

where $\mathcal{H}$ denotes the reproducing kernel Hilbert space associated to the chosen kernel. These two representations of the MMD allow to leverage the wide set of tools from both kernel methods and integral probability metric theories (see Sriperumbudur et al. 2009, 2012b for examples of the latter).

**Empirical estimator** Given a collection of samples $X = \{X_1, ..., X_n\}$ with empirical distribution $\hat{\rho}_n = \frac{1}{n} \sum_{1 \leq i \leq n} \delta(X_i)$, where $\delta(\cdot)$ denotes the Dirac delta, a natural empirical estimator of (1) is given by

$$\hat{\mu} := \mu(\hat{\rho}_n) = \frac{1}{n} \sum_{i=1}^{n} \phi(X_i) \tag{2}$$

which approximates the true kernel mean embedding at the rate $O(1/\sqrt{n})$ in $\|\cdot\|$ norm as discussed in Section 2.2. The time complexity for this evaluation grows linearly with the number $n$ of samples in the dataset. When the feature map $\phi$ is finite-dimensional, this estimator can be explicitly computed and stored, which might be reasonable when the feature map can efficiently be computed. However, when this is not the case, the whole dataset $X$ must be stored in memory, which can not only become prohibitive when the amount of available memory is limited, but also makes the manipulation of such embeddings unpractical. For instance, computing the maximum mean discrepancy between two empirical estimators computed from $n$ samples has a $\Theta(n^2)$ cost, which is not reasonable in applications where such distances must be computed repeatedly.

### 2.1. Problem

In this paper, we consider the problem of designing a new estimator $\hat{\mu}_m$ computed from $m$ points $\tilde{X}_1, ..., \tilde{X}_m$, which

1. can be computed more efficiently than $\hat{\mu}$;

2. preserves the statistical accuracy of $\hat{\mu}$.

The first requirement implies that $m$ should be smaller than the actual sample size $n$ while the second requirement can be expressed more formally as

$$\|\mu - \hat{\mu}_m\| = O(\|\mu - \hat{\mu}\|), \qquad (3)$$

provided that $m$ is large enough. Those two requirements induce a trade-off between *statistical accuracy* and *computational efficiency* of the desired estimator. We consider in particular the setting where the landmark points $(\tilde{X}_j)_{1 \le j \le m}$ are randomly sampled from the dataset $X$, so that the bound (3) must hold with high probability on the draw of these points.

A related problem that we tackle in Section 5 is that of the approximation of the maximum mean discrepancy. The approach that we explore consists in finding computationally-efficient approximations $\hat{\mu}_m(\rho_1), \hat{\mu}_m(\rho_2)$ of the kernel mean embeddings such that it holds with high probability

$$\|\hat{\mu}_m(\rho_1) - \hat{\mu}_m(\rho_2)\| = O(\|\mu(\rho_1) - \mu(\rho_2)\|).$$

## 2.2. Related work

The kernel mean embedding (1) can be approximated via its empirical counterpart (2) given $n$ samples from the distribution $\rho$, yielding an approximation error decreasing in $O(n^{-1/2})$. *Without any assumption* on the probability distribution $\rho$, the empirical average is asymptotically efficient and minimax optimal (Van der Vaart 2000, Theorem 25.21, Example 25.24) (see also (Lopez-Paz et al. 2015, Theorem 4)) thus there is no hope to get a better rate without any extra assumption on $\rho$. Tolstikhin et al. (2017) also showed that the minimax optimal rate for estimating kernel mean embeddings defined by continuous translation-invariant kernels on $\mathbb{R}^d$ is of order $O(n^{-1/2})$ for discrete measure and measures with infinitely differentiable densities (e.g., Gaussian and exponential). Approximation of the maximum mean discrepancy using empirical embeddings has also been studied by Sriperumbudur et al. (2012a).

Muandet et al. (2014) introduced shrinkage estimators of the form $\theta f + (1-\theta)\hat{\mu}$, where $\theta \in (0,1)$, $f \in \mathcal{H}$ is independent of the data and showed that those estimators perform better than the empirical average under mild assumptions on the probability distribution of interest. Such shrinkage strategies are complementary to our approach in the sense that they can be combined with any estimator of $\hat{\mu}$. In a different direction, Lerasle et al. (2019) proposed an outlier-robust estimator for the kernel mean embedding based on the median-of-means technique.

The principal way to efficiently approximate a kernel mean embedding is to consider an estimator of the form

$$\sum_{1 \le j \le m} \alpha_j \phi(y_j) \qquad (4)$$

with $m \ll n$. Multiple strategies exist to choose the points $(y_j)_{1 \le j \le m}$. For instance Cortes et al. (2014) proposed a greedy heuristic based on a notion of incoherence. Kernel herding (Y. Chen et al. 2010) is another deterministic way to select the points, which can be interpreted as an instance of the Frank-Wolfe algorithm (Bach et al. 2012) and has also connections with generalizations of orthogonal matching pursuit used in compressive sensing (Keriven et al. 2017). Depending on the context, the weights $(\alpha_j)_{1 \le j \le m}$ might be chosen uniform or directly optimized to minimize the approximation error or another related criterion. In a slightly different setting, Grünewälder et al. (2012) derive an equivalence between conditional mean embeddings and vector-valued regressors which allows them to use vector-valued regression methods to learn such embeddings.

These techniques are closely related to the quadrature problem, which consists in finding points $(y_j)_{1 \le j \le m}$ and weights $(\alpha_j)_{1 \le j \le m}$ approximating the integral

$$\int f(x)\,\mathrm{d}\rho(x) \approx \sum_{1 \le j \le m} \alpha_j f(y_j).$$

In the context of kernel methods, one typically wants to find an approximation scheme minimizing this error for all functions $f$ belonging to a reproducing kernel Hilbert space $\mathcal{H}$. In this case the approximation should not depend on the function $f$, and we have for any $f \in \mathcal{H}$ such that $\|f\| \le 1$ via the reproducing property

$$
\begin{aligned}
\text{QuadratureError} &:= \left| \int f(x)\,\mathrm{d}\rho(x) - \sum_{1 \le j \le m} \alpha_j f(y_j) \right| \\
&= \left| \left\langle f, \mu - \sum_{1 \le j \le m} \alpha_j \phi(y_j) \right\rangle_{\mathcal{H}} \right| \\
&\le \left\| \mu - \sum_{1 \le j \le m} \alpha_j \phi(y_j) \right\|.
\end{aligned}
$$

where $\mu$ is the mean embedding associated with $\rho$. This simple manipulation shows that approximate kernel mean embeddings are a way to design quadratures when the function $f$ to integrate is taken in a reproducing Hilbert space. Kernel quadrature bounds have in particular been obtained when sampling randomly and i.i.d the points $(y_j)_{1 \le j \le m}$ (Bach 2017), which includes approximation in the reproducing kernel Hilbert space as a special case. The sampling procedure in this work is however based on leverage scores, which are not exactly computable. Although algorithms exist to approximate these scores, the existing analysis does not cover this setting and further work would be needed in this direction. Bounds on the quadrature error have also been obtained for points sampled according to a determinantal point process (Belhadji et al. 2019), however sampling from such processes is also an expensive operation.

More generally, one can also consider mean embeddings associated to a finite-dimensional feature map $\phi_m$ approximating the kernel in the sense that $\langle \phi_m(x), \phi_m(y) \rangle \approx \kappa(x, y)$ for any $x, y$. One example would be to build $\phi_m$ using random features (Liu et al. 2021; Rahimi et al. 2008). This setting is particularly interesting from an algorithmic perspective, as the mean embedding can then be computed online and efficiently stored. As such representations lie in a different space, it is in general not possible to relate them to the original mean embedding. Yet, they remain a highly practical way to manipulate distributions, and the distance between such embeddings can still be interpreted as an estimator of the mean max discrepancy. However, because they are generic these methods might not adapt well to the data at hand, and we expect that our Nyström estimator might lead to a lower error for an equivalent complexity.

## 3. The Nyström estimator

As suggested above, our strategy to design an efficient algorithm is to consider an approximation of the form (4). More precisely, in our approach the points $(y_j)_{1 \leq j \leq m}$ on which the approximation is built are sampled uniformly and independently from the dataset. For this reason we denote them in the sequel $\tilde{X}_1, \ldots, \tilde{X}_m$, and our estimator will thus be in the space $\mathcal{H}_m := \operatorname{span}\left\{ \phi(\tilde{X}_1), \ldots, \phi(\tilde{X}_m) \right\}$ spanned by their features. Denoting $P_m$ the orthogonal projection on this subspace, the best estimator of the kernel mean embedding $\mu$ (1) in this space is $P_m \mu$. As this quantity can obviously not be computed when the distribution $\rho$ is unknown, we use instead

$$\hat{\mu}_m := P_m \hat{\mu}.$$

**Computation** It turns out that the weights of this embedding can be computed exactly. Denoting $K_m \in \mathbb{R}^{m \times m}$ and $K_{mn} \in \mathbb{R}^{m \times n}$ be the partial kernel matrices with entries $(K_m)_{ij} = \langle \phi(\tilde{X}_i), \phi(\tilde{X}_j) \rangle_{\mathcal{H}}$ for any $1 \leq i, j \leq m$ and $(K_{mn})_{ij} = \langle \phi(\tilde{X}_i), \phi(x_j) \rangle_{\mathcal{H}}$ for any $1 \leq i \leq m$ and $1 \leq j \leq n$, one can indeed check (see Appendix B) that our Nyström kernel mean embed-

ding estimator can be expressed as

$$\hat{\mu}_m = \sum_{1 \leq j \leq m} \alpha_j \phi(\tilde{X}_j) \quad \text{with} \quad \alpha = \frac{1}{n} K_m^+ K_{mn} 1_n, \quad (5)$$

where $1_n$ denotes a $n$-dimensional vector of ones and $K_m^+$ the (Moore-Penrose) pseudo-inverse of $K_m$.

**Complexity** The space complexity of the method is $\Theta(m^2 + md)$ for storing $K_m$ and the landmarks. Note that $K_{mn}$ does not need to be stored as $K_{mn}1_n$ can be computed sequentially in $\Theta(m)$ space. The time complexity is $\Theta(nmc_\kappa + m^3)$ where $c_\kappa$ corresponds to the cost of a kernel evaluation. The first term corresponds to the computation of $K_{mn}1_n$ while the second correspond to computing the pseudo-inverse of $K_m$ (numerically stable algorithms can still be used instead, but the complexity will be of this order regardless). When $\mathcal{X} = \mathbb{R}^d$, we will most often have $c_\kappa = d$.

## 4. Theoretical analysis

### 4.1. Setting and notations

We consider a probability space $(\mathcal{X}, \mathcal{B}, \rho)$ where $\mathcal{X}$ is a locally compact second countable topological space, $\mathcal{B}$ the Borel $\sigma$-algebra and $\rho$ is the data probability distribution with support in $\mathcal{X}$.

We define the (uncentered) covariance operator $C : \mathcal{H} \to \mathcal{H}$ as

$$C := \int \phi(x) \otimes_{\mathcal{H}} \phi(x) d\rho(x)$$

where $\phi(x) \otimes_{\mathcal{H}} \phi(x)$ denotes the rank one operator

$$(\phi(x) \otimes_{\mathcal{H}} \phi(x))(f) = \langle f, \phi(x) \rangle_{\mathcal{H}} \phi(x),$$

and denote by $C_\lambda = C + \lambda I$ its regularized version, for $\lambda > 0$. We now define, for any $\lambda > 0$ the functions

$$\forall x \in \mathcal{X}, \ \mathcal{N}_x(\lambda) := \langle \phi(x), C_\lambda^{-1} \phi(x) \rangle_{\mathcal{H}} \quad (6a)$$

$$\mathcal{N}(\lambda) := \mathbf{E}_x \mathcal{N}_x(\lambda) = \operatorname{tr}(CC_\lambda^{-1}) \quad (6b)$$

$$\mathcal{N}_\infty(\lambda) := \sup_{x \in \mathcal{X}} \mathcal{N}_x(\lambda). \quad (6c)$$

The quantity $\mathcal{N}(\lambda)$ is known as the effective dimension, and is a measure of the interaction between the kernel (or feature map) and the data probability distribution. It is tightly linked to the notion of leverage scores and has been shown to constitute a proper measure of hardness of kernel ridge regression problems (Alaoui et al. 2015). Finally, we denote by $\mathcal{L}(\mathcal{H})$ the set of bounded linear operators on $\mathcal{H}$ endowed with the operator norm $\|\cdot\|_{\mathcal{L}(\mathcal{H})}$.

The main assumption we make concerns the boundedness of the feature map.

**Assumption 4.1.** *There exists a positive constant $K < \infty$ such that $\sup_{x \in \mathcal{X}} \|\phi(x)\| \leq K$.*

This assumption ensures in particular that $\phi$ is integrable for any probability distribution over $\mathcal{X}$, and thus the kernel mean embedding (1) is well defined, interpreting the integral in (1) as a Bochner integral (Diestel et al. 1977, Chapter 2). Furthermore, it implies the two inequalities $\mathcal{N}_\infty(\lambda) \leq K^2/\lambda < \infty$ for any $\lambda > 0$, the latter being a crucial parameter to control the error of Nyström sampling. Finally, Assumption 4.1 implies that the operator C is a positive trace class operator on $\mathcal{H}$ and allows to leverage tools from spectral theory. We refer the reader to Rudi et al. (2015, Assumption 3) for a more detailed discussion around a similar assumption.

Assumption 4.1 is satisfied for feature maps derived from a large class of standard kernels such as, e.g., Gaussian and Laplacian radial basis functions kernels on the Euclidean space $\mathbb{R}^d$. It is also satisfied for polynomial kernels on a bounded space $\mathcal{X}$.

### 4.2. Error decomposition

In order to break down the approximation error, we introduce the quantity

$$\tilde{\mu}_m = \frac{1}{m} \sum_{j=1}^{m} \phi(\tilde{X}_j),$$

which is an unbiased estimate of the empirical kernel mean embedding $\hat{\mu}$. Note that, by definition, $\tilde{\mu}_m$ is an element of $\mathcal{H}_m$ and it corresponds to choosing uniform weights to average the landmarks. Furthermore, we introduce the orthogonal projection operator on the orthogonal of $\mathcal{H}_m$, $P_m^\perp := I - P_m$.

Our main result relies on the following deterministic error decomposition.

**Lemma 4.1 (Error decomposition).** *For any $\lambda > 0$, it holds (almost surely)*

$$\|\mu - \hat{\mu}_m\| \leq \|\mu - \hat{\mu}\| + \|P_m^\perp C_\lambda^{1/2}\|_{\mathcal{L}(\mathcal{H})} \|C_\lambda^{-1/2}(\hat{\mu} - \tilde{\mu}_m)\|.$$

*Proof of Lemma 4.1:* We use the decomposition

$$\|\mu - \hat{\mu}_m\| \leq \|\mu - \hat{\mu}\| + \|\hat{\mu} - \hat{\mu}_m\|$$

Note that

$$\|\hat{\mu} - \hat{\mu}_m\| = \|(I - P_m)\hat{\mu}\| = \|P_m^\perp(\hat{\mu} - \tilde{\mu}_m)\|$$

where the last inequality follows from $P_m^\perp \tilde{\mu}_m = 0$.

Hence we get

$$\|\mu - \hat{\mu}_m\| \leq \|\mu - \hat{\mu}\| + \|P_m^\perp(\hat{\mu} - \tilde{\mu}_m)\|$$
$$\leq \|\mu - \hat{\mu}\| + \|P_m^\perp C_\lambda^{1/2}\|_{\mathcal{L}(\mathcal{H})} \|C_\lambda^{-1/2}(\hat{\mu} - \tilde{\mu}_m)\|$$

which concludes the proof. □

Our strategy to bound the error $\|\mu - \hat{\mu}_m\|$ essentially consists in bounding those three terms separately using probabilistic concentration results in Hilbert spaces.

### 4.3. Main result

We now state our main result, and then specialize this result using additional knowledge on the spectral properties of the covariance operator.

**Theorem 4.1.** *Let Assumption 4.1 hold, and $m \geq 4$. Furthermore, assume that the data points $X_1, \dots, X_n$ are drawn i.i.d. from the distribution $\rho$ and that $m \leq n$ sub-samples $\tilde{X}_1, \dots, \tilde{X}_m$ are drawn uniformly with replacement from the dataset $\{X_1 \dots, X_n\}$. Then, it holds with probability at least $1 - \delta$ that*

$$\|\mu - \hat{\mu}_m\| \leq \frac{c_1}{\sqrt{n}} + \frac{c_2}{m}$$
$$+ \frac{c_3 \sqrt{\log(m/\delta)}}{m} \sqrt{\mathcal{N}\left(\frac{12K^2 \log(m/\delta)}{m}\right)},$$

*provided that*

$$m \geq \max(67, 12K^2 \|C\|_{\mathcal{L}(\mathcal{H})}^{-1}) \log\left(\frac{m}{\delta}\right),$$

*where $c_1, c_2, c_3$ are constants (made explicit in the proof) of order $K \log(1/\delta)$.*

A few remarks are in order regarding Theorem 4.1. First of all, denoting by $W$ the smallest branch of the Lambert's $W$ function on $]-e^{-1}, 0[$ (Weisstein 2002), the condition on the sub-sample size $m$ can also be expressed as $m \geq -W(-\delta/c)c$ with $c = \max(67, 12K^2 \|C\|_{\mathcal{L}(\mathcal{H})}^{-1})$ and can easily be checked on a computer.

Then, the bound on the error is split in three parts: the first part corresponds to the usual rate one gets estimating the kernel mean embedding by its standard empirical counterpart (as discussed in Section 2.2) while the second part and the third part result from our Nyström approximation scheme. Note that the first two terms already illustrate (at least partially) the trade-off between computational cost and statistical performance of our estimator: a small value of $m$ (i.e $m < \sqrt{n}$), while lightening the computational burden, yields a worst rate than the $O(1/\sqrt{n})$ rate; a large value of $m > \sqrt{n}$ does not improve the error rate, as it is governed by the first term in this case, and requires more computational and storage resources. This is expected since our estimator does not use more information on the distribution than the empirical estimator from which it is derived.

The precise trade-off should be settled by the third term. However, in its current form, it depends simultaneously on the subsample size $m$ and on the effective dimension $\mathcal{N}$. Extra assumptions about the effective dimension – i.e., about the interaction between the kernel and the probability distribution – are needed to obtain a more explicit and workable bound. To get a clearer picture, we derive sufficient conditions in the following corollary to get a $O(n^{-1/2})$ rate under various assumptions on the decay of the effective dimension.

**Corollary 4.1.** *Let the assumptions of Theorem 4.1 hold. Assume furthermore that the effective dimension $\mathcal{N}$ satisfies, for some $c > 0$,*

- *either $\mathcal{N}(\lambda) \leq c\lambda^{-\gamma}$ for some $\gamma \in ]0, 1]$,*

- *or $\mathcal{N}(\lambda) \leq \log(1 + c/\lambda)/\beta$, for some $\beta > 0$.*

*Then, choosing the subsample size $m$ to be*

- *$m = n^{1/(2-\gamma)} \log(n/\delta)$ in the first case;*

- *or $m = \sqrt{n} \log(\sqrt{n} \max(1/\delta, c/(6K^2)))$ in the second case,*

*we get*

$$\|\mu - \hat{\mu}_m\| = O\left(\frac{1}{\sqrt{n}}\right).$$

Let us comment on the assumptions on the decay of the effective dimension. Those assumptions can be linked to the decay of the eigenvalues $(\sigma_i)_i$ of the covariance matrix. First, note that the $\lambda^{-\gamma}$ rate always holds for $\gamma = 1$. More generally, it holds when the eigenvalues decay polynomially as $\sigma_i \lesssim i^{-1/\gamma}$, see e.g. Della Vecchia et al. (2021, Proposition 4). Similarly, the second rate holds for instance when the eigenvalues decay exponentially as $\sigma_i \leq ce^{-\lambda i}$, see e.g. Della Vecchia et al. (2021, Proposition 5).

Corollary 4.1 shows that, under mild assumptions on the kernel and on the decay of the effective dimension, our estimator fulfills the goals set in Section 2.1: it preserves the $O(1/\sqrt{n})$ error rate of the empirical estimator while reducing computational and storage costs.

Finally, we stress that our result only holds for uniform subsampling. While other sampling strategies, and in particular sampling proportionally to the so-called leverage scores (Alaoui et al. 2015), are known to be more effective in practice (in the sense that the same error might be obtained with a comparatively smaller value of $m$), it is not clear if and how our proof can be adapted to this setting.

Note that the quantity $\|\mu - \hat{\mu}_m\|^2$ can also be bounded by $\frac{1}{n}\|K_n - K_{\text{nys}}\|$ where $K_n$ and $K_{\text{nys}}$ denote respectively the full kernel matrix and its Nyström approximation, and $\|\cdot\|$ the operator norm. However, the works providing bounds in this context such as Kumar et al. (2012) and Musco et al. (2017) work in a fixed design. For this reason, such results would not directly be applicable in our setting, and it does not seem at first sight that any of the results imply the other.

## 5. Application: efficient estimation of the maximum mean discrepancy

The maximum mean discrepancy being built from kernel mean embeddings, its empirical estimation suffers as well from the computational issues discussed above. In this section, we introduce a sample-efficient estimator of the maximum mean discrepancy based on the kernel mean embedding estimator introduced in (5), and derive a high-probability upper bound on its approximation error.

We begin by stating the formal setting of this section. Let $\rho_1$ and $\rho_2$ be two probability measures whose supports are in the same space $\mathcal{X}$ (but not necessarily identical). We assume that we are given $n_k \geq 1$ i.i.d. samples $X_1^k, \ldots, X_{n_k}^k$ from $\rho_k$, for $k \in \{1, 2\}$. Let $P_k$ denote the orthogonal projection onto $\mathcal{H}_k = \text{span}\{\tilde{X}_1^k, \ldots, \tilde{X}_{m_k}^k\}$ where the landmarks $\tilde{X}_1^k, \ldots, \tilde{X}_{m_k}^k$ are independently sampled from the empirical distribution $\hat{\rho}_k = \frac{1}{n_k} \sum_{i=1}^n \delta(X_i^k)$.

Since the maximum mean discrepancy between two probability distributions corresponds to the norm of the difference between the mean embeddings of those distributions, our Nyström mean embedding estimator from (5) yields a natural estimator for the MMD:

$$\widehat{\text{MMD}} := \|\hat{\mu}_{m_1}^{(\rho_1)} - \hat{\mu}_{m_2}^{(\rho_2)}\|,$$

where for $k \in \{1, 2\}$,

$$\hat{\mu}_{m_k}^{(\rho_k)} = P_k \hat{\mu}(\rho_k).$$

The next result provides a high-probability upper bound on the difference (measured w.r.t. the norm of the feature space) between the true maximum mean discrepancy and our estimator,

$$\text{Err}_{(n_k, m_k)_k} := |\text{MMD}(\rho_1, \rho_2) - \widehat{\text{MMD}}|.$$

**Theorem 5.1 (Estimation error for the approximation of the MMD).** *Let Assumption 4.1 hold. Furthermore, assume that for $k \in \{1, 2\}$, the data points $X_1^k, \dots, X_{n_k}^k$ are drawn i.i.d. from the distribution $\rho_k$ and that $m_k \leq n_k$ sub-samples $\tilde{X}_1^k, \dots, \tilde{X}_{m_k}^k$ are drawn uniformly with replacement from the dataset $\{X_1^k \dots, X_{n_k}^k\}$. Then, for any $\delta \in ]0, 1[$, it holds with probability at least $1 - 2\delta$*

$$
\begin{aligned}
Err_{(n_k, m_k)_k} \leq & \sum_{k \in \{1, 2\}} \frac{c_1}{\sqrt{n_k}} + \frac{c_2}{m_k} \\
& + \frac{\sqrt{\log(m_k/\delta)}}{m_k} \sqrt{\mathcal{N}^{(\rho_k)} \left( \frac{12K^2 \log(m_k/\delta)}{m_k} \right)},
\end{aligned}
$$

*provided that, for $k \in \{1, 2\}$,*

$$m_k \geq \max(67, 12K^2 \|C_k\|_{\mathcal{L}(\mathcal{H})}^{-1}) \log\left(\frac{m_k}{\delta}\right),$$

*where $c_1 = 2K\sqrt{2\log(6/\delta)}$, $c_2 = 4\sqrt{3}K\log(12/\delta)$ and $c_3 = 6K\sqrt{\log(12/\delta)}$. The notation $\mathcal{N}^{(\rho_k)}$ denotes the effective dimension associated to the distribution $\rho_k$.*

The high-probability upper bound provided in the above theorem is similar in spirit to that of Theorem 4.1. It follows by replicating the proof of Theorem 4.1 for each measure $\rho_1$ and $\rho_2$ separately. As in Corollary 4.1, the third term can be made more explicit by making assumptions on the effective dimensions $\mathcal{N}^{(\rho_k)}, k \in \{1, 2\}$ and setting the right subsample sizes $(m_k)_{k \in \{1, 2\}}$. For more flexibility we allowed the sample and subsample sizes to differ between the two probability distributions. In the case of large sample imbalance, it might be interesting to choose a larger subsample size for the smallest sample to improve the statistical error rate, since it is guided by the smallest (sub)sample sizes.

As a direction for future work, we note that, in a similar spirit to Gretton et al. (2012), it is possible to derive a two-sample test from our bound in Theorem 5.1.

Designing efficient MMD-based tests remains an important research direction, and the simplicity of our approach should allow to reduce computational costs compared to state-of-the-art algorithms (Jitkrittum et al. 2016). However, this requires further development that are outside the scope of this paper, and for this reason we do not compare our approach to works focusing on efficient testing in the following section.

# 6. Numerical experiments

In this section, we provide experimental results to illustrate our theoretical findings on the decay of the approximation error. We provide a proof of work in a simple experimental setting, but extending these results to broader families of datasets and kernel types would be interesting in the future.

## 6.1. Synthetic data

We first generate data according to a Gaussian mixture, i.e. we choose $\mathcal{X} = \mathbb{R}^d$ and $\rho = \sum_{1 \leq i \leq p} \frac{1}{p} \mathcal{N}(\mu_i, I)$ where the centers $(\mu_i)_{1 \leq i \leq p}$ are themselves drawn according to a multivariate normal distribution $\mathcal{N}(0, 5I)$. For our experiments, we choose $d = 10$, $p = 8$. We use the Gaussian kernel $k(x, y) = \exp(-\|x - y\|_2^2/(2\sigma_k^2))$, which satisfies Assumption 4.1.

To efficiently compute the approximation error of the embedding $\sum_{j=1}^m \alpha_i \phi(x_i)$, we use the decomposition

$$
\begin{aligned}
\left\| \mu - \sum_{j=1}^m \alpha_i \phi(x_i) \right\|^2 = & \iint k(x, y) \, d\rho(x) \, d\rho(y) + \alpha^T K_X \alpha \\
& - 2 \sum_{j=1}^m \alpha_i \int k(x, x_i) \, d\rho(x)
\end{aligned}
$$

which follows from the definition of $\mu$ (1), where $K_X$ denotes the kernel matrix of the $(x_j)_{1 \leq j \leq m}$ and $\alpha = [\alpha_1, \dots, \alpha_m]$. Both integrals in this expression can be computed in closed form for the considered distribution and kernel as detailed in Appendix G.

Figure 1 shows the obtained error for both the empirical estimator and the Nyström estimator when varying the size of the support $m$. Both estimators are computed using a sample of $n = \{10^3, 10^4, 10^5\}$ points drawn from $\rho$, and the standard deviation $\sigma_k$ of the kernel is chosen to be the median of the inter-points distance, estimated for efficiency on a random subset of 1000 points. In this setting the data distribution is sub-Gaussian, and as we are using a Gaussian kernel the spectrum of the covariance matrix decays exponentially (Widom 1963), i.e. we are in the second setting of Corollary 4.1 and we expect to need $m \geq \sqrt{n} \log(\sqrt{n})$ to achieve the same rate as the empirical estimator. This is indeed quite close to what is observed in practice. We recall that the time complexity of our method is $\Theta(nmd + m^3)$, and we also include in the settings where $n$ is moderate results for the greedy methods of Cortes et al. (2014) (which runs in $\Theta(nmd + m^3)$ and $\Theta(nm)$ space), L. Chen et al. (2018) ($\Theta(nm(m + d))$) and Keriven et al. (2017) ($\Theta(nd^2m + Id^2m^3)$) where $I$ is a fixed number of iterations, using a sketch of size $md$; although this method makes use of random features, we use the recovered landmarks to build an estimate in the reproducing kernel Hilbert space). Our method performs similarly to the first two, which come with no statistical approximation guarantees, while the method of Keriven et al. (2017) yields better approximations but at a significantly higher cost.

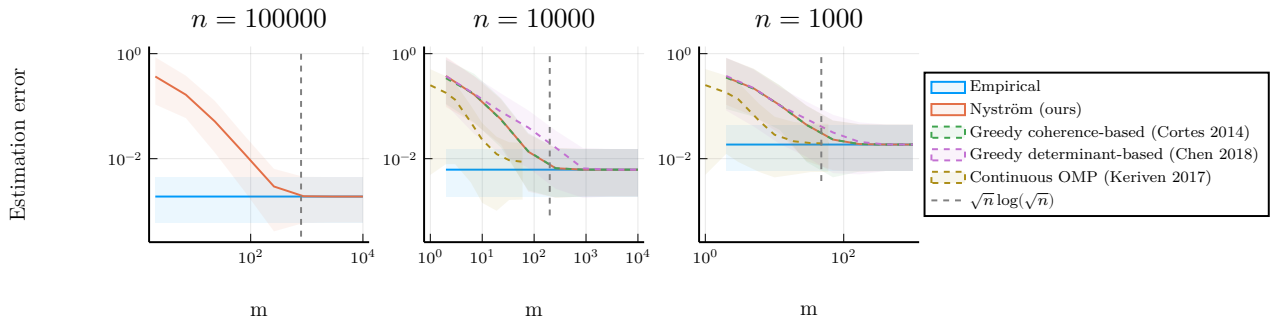

Figure 1: Estimation errors $\|\mu - \hat{\mu}\|$ (in blue) and $\|\mu - \hat{\mu}_m\|$ (in orange) as well as for other heuristics for synthetic Gaussian data. Means (solid lines), 5-percentiles and 95-percentiles over 100 trials (shaded regions). The data distribution is fixed and identical for the 3 values of $n$.

Although we only plot the error of the mean embedding estimation for conciseness, the error of the MMD estimation has exactly the same behavior.

### 6.2. Real datasets

We then perform experiments with data from the Fasttext[1] (Bojanowski et al. 2016) (english features), FMA[2] (Defferrard et al. 2016) (MFCC features), Intel Lab[3] and Gowalla[4] (Cho et al. 2011) datasets, which respectively consist of text features in dimension 300, audio features in dimension 20, physical measurements in dimension 5 and geolocation data in dimension 2. We use the whole dataset for FMA, resulting in $N = 106574$ points, while for Fasttext, Gowalla and Intel Lab we limit ourselves to $N = 100000$ randomly selected points. For each dataset, we consider $\rho$ to be the uniform distribution over these points, and we build the empirical estimator using a random sample of size $n = 10^4$.

Results are presented in Figure 2, and display the same behavior as with synthetic data. While we don't have any particular guarantees on the spectrum's decay in this setting, it is still verified empirically than choosing $m$ of the order of $\sqrt{n}\log(\sqrt{n})$ is sufficient to obtain an error of the same order as the empirical estimator for all datasets but Fasttext. Note that we did not optimize the kernel choice for all our experiments. It could be the case that an even faster decay of the error happens for an optimized choice of kernel. The slower error decrease for the Nyström estimator on the Fasttext dataset could be explained by many factors such as, for instance, a slow decay of the covariance operator eigenvalues. This issue requires further empirical exploration, which is out of scope for this paper. On the other side, on the Intel Lab dataset it seems that the error of the Nyström estimator decays even faster and only $m \approx 20$ features are sufficient to obtain the same error as the empirical estimator. Here again, similar results can be obtained for the maximum mean discrepancy but are omitted for conciseness.

## 7. Conclusion

In this article, we introduced a simple and efficient estimator of the kernel mean embedding based on the Nyström approximation. Both theoretical and empirical results show that in most settings, an approximation supported on the features of $m \ll n$ points (typically $m \approx \sqrt{n}$) will have the same error than that of the empirical estimator while enjoying much better computational properties and being efficiently storable.

An interesting question for future work is to see whether our results can be improved with other sampling strategies: while it is well known that leverage scores sampling allows in many related contexts to achieve the same error with a smaller number of landmarks, it is not clear for now how to obtain meaningful theoretical guarantees under this sampling scheme.

---

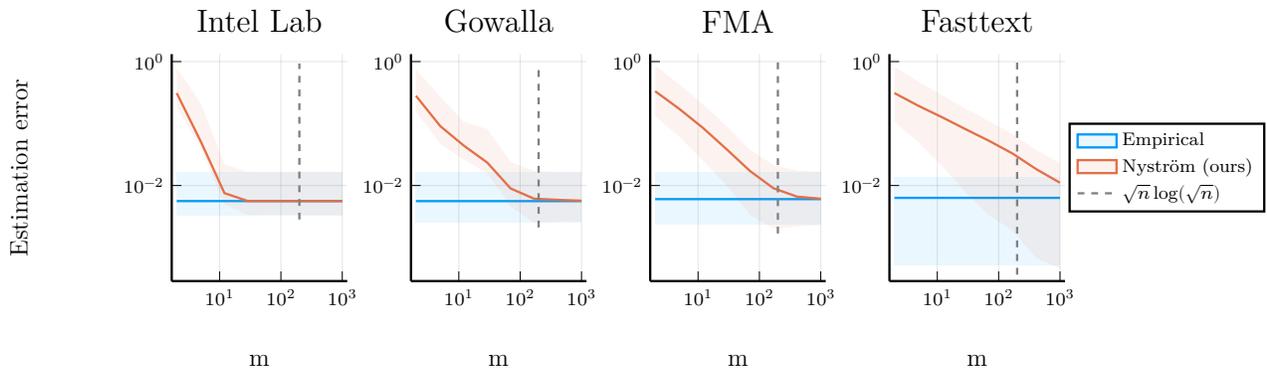

Figure 2: Estimation errors for the empirical estimator $\|\mu - \hat{\mu}\|$ (in blue) and the Nyström estimator $\|\mu - \hat{\mu}_m\|$ (in orange) for real datasets. Means (solid lines), 5-percentiles and 95-percentiles over 100 trials (shaded regions). The data distribution is fixed and identical for the 3 values of $n$.

## Acknowledgments

L.R. acknowledges the financial support of the European Research Council (grant SLING 819789), the AFOSR project FA9550-18-1-7009 (European Office of Aerospace Research and Development), the EU H2020-MSCA-RISE project NoMADS - DLV-777826, and the Center for Brains, Minds and Machines (CBMM), funded by NSF STC award CCF-1231216. We thank the anonymous referees for their helpful comments and suggestions.

## References

Alaoui, Ahmed and Michael W. Mahoney (2015). "Fast Randomized Kernel Ridge Regression with Statistical Guarantees". In: *Advances in Neural Information Processing Systems*, pp. 775–783.

Bach, Francis (2017). "On the Equivalence between Kernel Quadrature Rules and Random Feature Expansions". In: *The Journal of Machine Learning Research* 18.1, pp. 714–751.

Bach, Francis, Simon Lacoste-Julien, and Guillaume Obozinski (June 26, 2012). "On the Equivalence between Herding and Conditional Gradient Algorithms". In: *Proceedings of the 29th International Coference on International Conference on Machine Learning*. ICML'12. Madison, WI, USA: Omnipress, pp. 1355–1362. ISBN: 978-1-4503-1285-1.

Balog, Matej, Ilya Tolstikhin, and Bernhard Schölkopf (Oct. 2018). "Differentially Private Database Release via Kernel Mean Embeddings". In: *Proceedings of the 35th International Conference on Machine Learning*. Ed. by Jennifer Dy and Andreas Krause. Vol. 80. Proceedings of Machine Learning Research. PMLR, pp. 414–422.

Belhadji, Ayoub, Rémi Bardenet, and Pierre Chainais (Dec. 31, 2019). "Kernel Quadrature with DPPs". In: Advances in Neural Information Processing Systems. Vol. 32, pp. 12927–12937. arXiv: 1906.07832 [cs, stat].

Bojanowski, Piotr, Edouard Grave, Armand Joulin, and Tomas Mikolov (2016). "Enriching Word Vectors with Subword Information". arXiv: 1607.04606.

Borgwardt, Karsten M., Arthur Gretton, Malte J. Rasch, Hans-Peter Kriegel, Bernhard Schölkopf, and Alexander J. Smola (2006). "Integrating structured biological data by Kernel Maximum Mean Discrepancy". In: *Proceedings 14th International Conference on Intelligent Systems for Molecular Biology 2006, Fortaleza, Brazil, August 6-10, 2006*, pp. 49–57.

Chatalic, Antoine, Vincent Schellekens, Florimond Houssiau, Yves-Alexandre De Montjoye, Laurent Jacques, and Rémi Gribonval (May 15, 2021). "Compressive Learning with Privacy Guarantees". In: *Information and Inference: A Journal of the IMA* (iaab005). ISSN: 2049-8772.

Chen, Laming, Guoxin Zhang, and Eric Zhou (2018). "Fast Greedy MAP Inference for Determinantal Point Process to Improve Recommendation Diversity". In: *Advances in Neural Information Processing Systems* 31, pp. 5627–5638.

Chen, Yutian, Max Welling, and Alex Smola (July 8, 2010). "Super-Samples from Kernel Herding". In: *Proceedings of the Twenty-Sixth Conference on Uncertainty in Artificial Intelligence*. UAI'10. Arlington, Virginia, USA: AUAI Press, pp. 109–116. ISBN: 978-0-9749039-6-5.

Cho, Eunjoon, Seth A. Myers, and Jure Leskovec (2011). "Friendship and Mobility: User Movement in Location-Based Social Networks". In: *Proceedings* of the 17th ACM SIGKDD International Conference on Knowledge Discovery and Data Mining, pp. 1082–1090.

Cortes, Efren Cruz and Clayton Scott (May 2014). "Scalable Sparse Approximation of a Sample Mean". In: *2014 IEEE International Conference on Acoustics, Speech and Signal Processing (ICASSP)*. 2014 IEEE International Conference on Acoustics, Speech and Signal Processing (ICASSP). Florence, Italy: IEEE, pp. 5237–5241. ISBN: 978-1-4799-2893-4.

Defferrard, Michaël, Kirell Benzi, Pierre Vandergheynst, and Xavier Bresson (2016). "FMA: A Dataset for Music Analysis". In: *ISMIR*. arXiv: 1612.01840.

Della Vecchia, Andrea, Jaouad Mourtada, Ernesto De Vito, and Lorenzo Rosasco (Feb. 25, 2021). "Regularized ERM on Random Subspaces". arXiv: 2006.10016 [cs, stat].

Diestel, Joseph and John Jerry Uhl (1977). *Vector Measures*. Mathematical Surveys and Monographs 15. American Mathematical Soc. ISBN: 0-8218-1515-6 978-0-8218-1515-1.

Gretton, Arthur, Karsten M. Borgwardt, Malte J. Rasch, Bernhard Schölkopf, and Alexander J. Smola (2012). "A Kernel Two-Sample Test". In: *J. Mach. Learn. Res.* 13, pp. 723–773.

Grünewälder, Steffen, Guy Lever, Luca Baldassarre, Sam Patterson, Arthur Gretton, and Massimilano Pontil (July 24, 2012). "Conditional Mean Embeddings as Regressors - Supplementary". arXiv: 1205.4656 [cs, stat].

Hayati, Saeed, Kenji Fukumizu, and Afshin Parvardeh (2020). "Kernel Mean Embedding of Probability Measures and its Applications to Functional Data Analysis". In: *arXiv preprint arXiv:2011.02315*.

Jitkrittum, Wittawat, Zoltán Szabó, Kacper P Chwialkowski, and Arthur Gretton (2016). "Interpretable Distribution Features with Maximum Testing Power". In: *Advances in Neural Information Processing Systems*. Vol. 29. Curran Associates, Inc.

Keriven, Nicolas, Nicolas Tremblay, Yann Traonmilin, and Rémi Gribonval (Mar. 5, 2017). "Compressive K-means". In: International Conference on Acoustics, Speech and Signal Processing (ICASSP).

Kim, Kee-Eung and Hyun Soo Park (2018). "Imitation Learning via Kernel Mean Embedding". In: *Proceedings of the Thirty-Second AAAI Conference on Artificial Intelligence, (AAAI-18), the 30th innovative Applications of Artificial Intelligence (IAAI-18), and the 8th AAAI Symposium on Educational Advances in Artificial Intelligence (EAAI-18), New Orleans, Louisiana, USA, February 2-7, 2018*. Ed. by Sheila A. McIlraith and Kilian Q. Weinberger. AAAI Press, pp. 3415–3422.

Kumar, Sanjiv, Mehryar Mohri, and Ameet Talwalkar (2012). "Sampling Methods for the Nyström Method". In: *The Journal of Machine Learning Research* 13.1, pp. 981–1006.

Laub, Alan J. (2004). *Matrix Analysis for Scientists and Engineers*. SIAM: Society for Industrial and Applied Mathematics. ISBN: 978-0-89871-576-7.

Lerasle, Matthieu, Zoltán Szabó, Timothée Mathieu, and Guillaume Lecué (2019). "MONK Outlier-Robust Mean Embedding Estimation by Median-of-Means". In: *Proceedings of the 36th International Conference on Machine Learning, ICML 2019, 9-15 June 2019, Long Beach, California, USA*. Ed. by Kamalika Chaudhuri and Ruslan Salakhutdinov. Vol. 97. Proceedings of Machine Learning Research. PMLR, pp. 3782–3793.

Li, Chun-Liang, Wei-Cheng Chang, Yu Cheng, Yiming Yang, and Barnabás Póczos (2017). "MMD GAN: Towards Deeper Understanding of Moment Matching Network". In: *Advances in Neural Information Processing Systems 30: Annual Conference on Neural Information Processing Systems 2017, December 4-9, 2017, Long Beach, CA, USA*. Ed. by Isabelle Guyon, Ulrike von Luxburg, Samy Bengio, Hanna M. Wallach, Rob Fergus, S. V. N. Vishwanathan, and Roman Garnett, pp. 2203–2213.

Liu, Fanghui, Xiaolin Huang, Yudong Chen, and Johan A. K. Suykens (Mar. 16, 2021). "Random Features for Kernel Approximation: A Survey on Algorithms, Theory, and Beyond". arXiv: 2004.11154 [cs, stat].

Lopez-Paz, David, Krikamol Muandet, Bernhard Scholkopf, Ilya Tolstikhin, and Lopezpaz Org (2015). "Towards a Learning Theory of Cause-Effect Inference". In: *Proceedings of the 32nd International Conference on Machine Learning*. Vol. PMLR 37, pp. 1452–1461.

Muandet, Krikamol, Kenji Fukumizu, Francesco Dinuzzo, and Bernhard Schölkopf (2012). "Learning from Distributions via Support Measure Machines". In: *Advances in Neural Information Processing Systems 25: 26th Annual Conference on Neural Information Processing Systems 2012. Proceedings of a meeting held December 3-6, 2012, Lake Tahoe, Nevada, United States*. Ed. by Peter L. Bartlett, Fernando C. N. Pereira, Christopher J. C. Burges, Léon Bottou, and Kilian Q. Weinberger, pp. 10–18.

Muandet, Krikamol, Kenji Fukumizu, Bharath Sriperumbudur, Arthur Gretton, and Bernhard Schoelkopf (Jan. 27, 2014). "Kernel Mean Estimation and Stein Effect". In: *Proceedings of the 31st International Conference on Machine Learning*. International Conference on Machine Learning. PMLR, pp. 10–18.

Muandet, Krikamol, Kenji Fukumizu, Bharath K. Sriperumbudur, and Bernhard Schölkopf (2017). "Kernel Mean Embedding of Distributions: A Review and Beyond". In: *Found. Trends Mach. Learn.* 10.1-2, pp. 1–141.

Müller, Alfred (1997). "Integral probability metrics and their generating classes of functions". In: *Advances in Applied Probability* 29.2, pp. 429–443.

Musco, Cameron and Christopher Musco (2017). "Recursive Sampling for the Nystrom Method". In: *Advances in Neural Information Processing Systems*, pp. 3833–3845.

Park, Mijung, Wittawat Jitkrittum, and Dino Sejdinovic (2016). "K2-ABC: Approximate Bayesian Computation with Kernel Embeddings". In: *Proceedings of the 19th International Conference on Artificial Intelligence and Statistics, AISTATS 2016, Cadiz, Spain, May 9-11, 2016.* Ed. by Arthur Gretton and Christian C. Robert. Vol. 51. JMLR Workshop and Conference Proceedings. JMLR.org, pp. 398–407.

Petersen, Kaare Brandt, Michael Syskind Pedersen, et al. (2008). "The matrix cookbook". In: *Technical University of Denmark* 7.15, p. 510.

Pinelis, Iosif (1994). "Optimum Bounds for the Distributions of Martingales in Banach Spaces". In: *The Annals of Probability*, pp. 1679–1706.

Rahimi, Ali and Benjamin Recht (2008). "Random Features for Large-Scale Kernel Machines". In: *Advances in Neural Information Processing Systems*, pp. 1177–1184.

Rudi, Alessandro, Raffaello Camoriano, and Lorenzo Rosasco (2015). "Less Is More: Nyström Computational Regularization". In: *Proceedings of the 28th International Conference on Neural Information Processing Systems - Volume 1.* NIPS'15. Cambridge, MA, USA: MIT Press, pp. 1657–1665.

Shahrampour, Shahin and Soheil Kolouri (2019). "On sampling random features from empirical leverage scores: Implementation and theoretical guarantees". In: *arXiv preprint arXiv:1903.08329.*

Smola, Alexander J., Arthur Gretton, Le Song, and Bernhard Schölkopf (2007). "A Hilbert Space Embedding for Distributions". In: *Algorithmic Learning Theory, 18th International Conference, ALT 2007, Sendai, Japan, October 1-4, 2007, Proceedings.* Ed. by Marcus Hutter, Rocco A. Servedio, and Eiji Takimoto. Vol. 4754. Lecture Notes in Computer Science. Springer, pp. 13–31.

Song, Le, Kenji Fukumizu, and Arthur Gretton (2013). "Kernel Embeddings of Conditional Distributions: A Unified Kernel Framework for Nonparametric Inference in Graphical Models". In: *IEEE Signal Process. Mag.* 30.4, pp. 98–111.

Sriperumbudur, Bharath K., Kenji Fukumizu, Arthur Gretton, Bernhard Schölkopf, and Gert RG Lanckriet (2009). "On integral probability metrics,\phi-divergences and binary classification". In: *arXiv preprint arXiv:0901.2698.*

– (2012a). "On the Empirical Estimation of Integral Probability Metrics". In: *Electronic Journal of Statistics* 6, pp. 1550–1599.

– (2012b). "On the empirical estimation of integral probability metrics". In: *Electronic Journal of Statistics* 6, pp. 1550–1599.

Sriperumbudur, Bharath K., Arthur Gretton, Kenji Fukumizu, Bernhard Schölkopf, and Gert R. G. Lanckriet (2010). "Hilbert Space Embeddings and Metrics on Probability Measures". In: *Journal of Machine Learning Research* 11 (Apr), pp. 1517–1561. ISSN: ISSN 1533-7928.

Sutherland, Danica J., Hsiao-Yu Tung, Heiko Strathmann, Soumyajit De, Aaditya Ramdas, Alexander J. Smola, and Arthur Gretton (2017). "Generative Models and Model Criticism via Optimized Maximum Mean Discrepancy". In: *5th International Conference on Learning Representations, ICLR 2017, Toulon, France, April 24-26, 2017, Conference Track Proceedings.* OpenReview.net.

Tolstikhin, Ilya, Bharath K Sriperumbudur, and Krikamol Muandet (2017). "Minimax Estimation of Kernel Mean Embeddings". In: *The Journal of Machine Learning Research* 18.1, pp. 3002–3048.

Van der Vaart, Aad W. (2000). *Asymptotic Statistics.* Vol. 3. Cambridge university press.

Weisstein, Eric W (2002). "Lambert W-function". In: *https://mathworld. wolfram. com/.*

Widom, Harold (1963). "Asymptotic Behavior of the Eigenvalues of Certain Integral Equations". In: *Transactions of the American Mathematical Society* 109.2, pp. 278–295.

Yurinsky, Vadim (1995). *Sums and Gaussian Vectors.* 1st ed. Lecture Notes in Mathematics 1617. Springer-Verlag Berlin Heidelberg. ISBN: 978-3-540-60311-5.

Zhang, Kun, Bernhard Schölkopf, Krikamol Muandet, and Zhikun Wang (2013). "Domain Adaptation under Target and Conditional Shift". In: *Proceedings of the 30th International Conference on Machine Learning, ICML 2013, Atlanta, GA, USA, 16-21 June 2013.* Vol. 28. JMLR Workshop and Conference Proceedings. JMLR.org, pp. 819–827.

Zou, Shaofeng, Yingbin Liang, H. Vincent Poor, and Xinghua Shi (2014). "Nonparametric Detection of Anomalous Data via Kernel Mean Embedding". In: *CoRR* abs/1405.2294. arXiv: 1405.2294.

## Structure of the Appendix

We begin by introducing additional notations in Appendix A. Then, in Appendix B, we prove how to obtain the claim made in Eq. (5). Appendix C contains the proof of Theorem 4.1 as well as its corollaries. Appendix D contains the proof of Theorem 5.1. We gather in Appendix E the concentration results that our proof of Theorem 4.1 rely on. We recall in Appendix F a key result for Nyström subsampling. Finally we detail some computations used in the numerical experiments in Appendix G.

## A. Additional notations

We define the empirical covariance operator as

$$\hat{C}_n = \sum_{i=1}^{n} \phi(x_i) \otimes_{\mathcal{H}} \phi(x_i).$$

For any operator $Q : \mathcal{H} :\rightarrow \mathcal{H}$ and any real number $\lambda > 0$, we denote by $Q_\lambda : \mathcal{H} \rightarrow \mathcal{H}$ the regularized operator $Q_\lambda = Q + \lambda I$. We denote the (Moore-Penrose) pseudo-inverse of an operator $A$ by $A^+$.

Given a random variable $X$, we write $\operatorname{ess\,sup} X$ to denote its essential supremum.

$1_n \in \mathbb{R}^n$ denotes the $n$-dimensional vector of ones.

## B. Derivation of the weights

This section provides a proof for the claim in Eq. (5). For ease of exposition, let us introduce the operators

$$\Phi_m : \mathbb{R}^m \rightarrow \mathcal{H}_m, \, \alpha \mapsto \sum_{j=1}^{m} \alpha_j \phi(\tilde{X}_j),$$

$$\Phi_n : \mathbb{R}^n \rightarrow \mathcal{H}, \, \alpha \mapsto \sum_{i=1}^{n} \alpha_i \phi(X_i).$$

Since, by definition, $\hat{\mu}_m$ is the orthogonal projection of $\hat{\mu}$ onto the space $\mathcal{H}_m$, it can be expressed as $\hat{\mu}_m = \Phi_m \alpha$ where the weights $\alpha \in \mathbb{R}^m$ minimize the mapping $\beta \mapsto \|\hat{\mu} - \Phi_m \beta\|^2$. Setting the gradient of this mapping to zero, we obtain that $\alpha$ must satisfy

$$\Phi_m^* \Phi_m \alpha = \Phi_m^* \hat{\mu}.$$

The minimum norm solution of the above equation is given by $\alpha = (\Phi_m^* \Phi_m)^+ \Phi_m^* \hat{\mu}$ (Laub 2004). Noting that the empirical measure $\hat{\mu}$ can be expressed as $\hat{\mu} = \frac{1}{n} \Phi_n 1_n$ and using the fact that $\Phi_m^* \Phi_m = K_m$, $\Phi_m^* \Phi_n = K_{mn}$, we obtain the claimed equality

$$\alpha = K_m^+ \Phi_m^* (n^{-1} \Phi_n 1_n) = \frac{1}{n} K_m^+ K_{mn} 1_n.$$

## C. Proof of the main result

### C.1. Generic bound

Theorem 4.1 is a consequence of a generic bound which we state now.

**Lemma C.1.** *Let Assumption 4.1 hold. Furthermore, assume that the data points $X_1, \dots, X_n$ are drawn i.i.d. from the distribution $\rho$ and that $m \leq n$ sub-samples $\tilde{X}_1, \dots, \tilde{X}_m$ are drawn uniformly with replacement from the dataset $\{X_1 \dots, X_n\}$. Then, for any $\lambda \in ]0, \|C\|_{\mathcal{L}(\mathcal{H})}]$ and $\delta \in ]0, 1[$, with probability at least $1 - \delta$*

$$\|\mu - \hat{\mu}_m\| \leq \frac{2K\sqrt{2\log(6/\delta)}}{\sqrt{n}} + \sqrt{\lambda} \left( \frac{4\sqrt{3\mathcal{N}_\infty(\lambda)}\log(12/\delta)}{m} + 6\sqrt{\frac{\mathcal{N}(\lambda)\log(12/\delta)}{m}} \right),$$

*provided that*

- $m \geq \max(67, 5\mathcal{N}_\infty(\lambda)) \log \frac{12K^2}{\lambda\delta}$,

- $\lambda n \geq 12K^2 \log(4/\delta)$.

*Proof of Lemma C.1:* Let $\delta \in (0,1)$ be the desired confidence level. Let $\lambda > 0$, $m \in \mathbb{N}$ and $n \in \mathbb{N}$ satisfy the conditions of the theorem. Using the error decomposition of Lemma 4.1, we get

$$\|\mu - \hat{\mu}_m\| \leq \|\mu - \hat{\mu}\| + \|P_m^\perp C_\lambda^{1/2}\|_{\mathcal{L}(\mathcal{H})} \|C_\lambda^{-1/2}(\hat{\mu} - \tilde{\mu}_m)\|.$$

Controlling the first term amounts to measuring the concentration of an empirical mean around its true mean in a Hilbert space. Multiple variants of such results can be found in the literature (see, *e.g.*, (Pinelis 1994)). We apply here Lemma E.1 on the random variables $\eta_i := \phi(X_i) - \mu, 1 \leq i \leq n$. Note that they are indeed bounded since, for any index $1 \leq i \leq n$, $\|\eta_i\| \leq 2\sup_{x \in \mathcal{X}} \|\phi(x)\| = 2K$. Thus, it holds with probability at least $1 - {}^\delta/_3$ on the draw of the the dataset $X_1, \ldots, X_n$ that

$$\|\mu - \hat{\mu}\| \leq \frac{2K\sqrt{2\log(6/\delta)}}{\sqrt{n}}.$$

Next, we rely on Lemma F.1 to bound the term $\|P_m^\perp C_\lambda^{1/2}\|_{\mathcal{L}(\mathcal{H})}$ with high probability. Since the Nyström landmarks are uniformly drawn and $m \geq \max(67, 5\mathcal{N}_\infty(\lambda)) \log \frac{12K^2}{\lambda\delta}$, we have, for any $\lambda > 0$, with probability at least $1 - {}^\delta/_3$ on the draw of the landmarks $\tilde{X}_1, \ldots, \tilde{X}_m$,

$$\|P_m^\perp C_\lambda^{1/2}\|_{\mathcal{L}(\mathcal{H})} \leq \sqrt{3\lambda}.$$

Finally, the last term can be bounded using Lemma E.5 which implies that, since $\lambda$ satisfies $0 < \lambda \leq \|C\|_{\mathcal{L}(\mathcal{H})}$ and $\lambda n \geq 12K^2 \log(4/\delta)$, it holds with probability at least $1 - {}^\delta/_3$

$$\left\|C_\lambda^{-1/2}(\hat{\mu} - \tilde{\mu}_m)\right\| \leq \frac{4\sqrt{\mathcal{N}_\infty(\lambda)}\log(4/\delta_3)}{m} + \sqrt{\frac{12\mathcal{N}(\lambda)\log(4/\delta_3)}{m}}.$$

Taking the union bound over the three events yields the desired result: with probability at least $1 - \delta$ (over all sources of randomness), it holds that

$$\|\mu - \hat{\mu}_m\| \leq \frac{2K\sqrt{2\log(6/\delta)}}{\sqrt{n}} + \sqrt{3\lambda}\left(\frac{4\sqrt{\mathcal{N}_\infty(\lambda)}\log(12/\delta)}{m} + \sqrt{\frac{12\mathcal{N}(\lambda)\log(12/\delta)}{m}}\right).$$

$\square$

## C.2. Proof of Theorem 4.1

*Proof of Theorem 4.1:* Assuming that our choice of $m$ and $\lambda$ satisfies the constraints

$$\begin{cases} m \geq \max(67, 5\mathcal{N}_\infty(\lambda)) \log \frac{12K^2}{\lambda\delta} \\ \lambda n \geq 12K^2 \log(4/\delta) \\ 0 < \lambda \leq \|C\|_{\mathcal{L}(\mathcal{H})} \end{cases}, \tag{7}$$

we can apply Lemma C.1 and use the fact that $\mathcal{N}_\infty(\lambda) \leq K^2/\lambda$ to get

$$\|\mu - \hat{\mu}_m\| \leq \frac{2K\sqrt{2\log(6/\delta)}}{\sqrt{n}} + \frac{4\sqrt{3}K\log(12/\delta)}{m} + 6\sqrt{\log(12/\delta)}\sqrt{\frac{\lambda\mathcal{N}(\lambda)}{m}}.$$

Setting $\lambda = \frac{12K^2 \log(m/\delta)}{m}$ we obtain the claimed result with constants $c_1 = 2K\sqrt{2\log(6/\delta)}$, $c_2 = 4\sqrt{3}K\log(12/\delta)$, and $c_3 = 12\sqrt{3\log(12/\delta)}K$.

Let us now check that our choices are consistent with the constraints. We will also obtain a more user-friendly expression for the constraints and express the sub-sample size $m$ as a function of the sample size $n$. Using the fact that $\mathcal{N}_\infty(\lambda) \leq K^2/\lambda$, one can easily check that a sufficient set of conditions to satisfy (7) is given by

$$\begin{cases} m \geq 67 \log\left(\frac{1}{\delta} \frac{m}{\log(m/\delta)}\right) \\ m \geq \frac{5m}{12 \log\left(\frac{m}{\delta}\right)} \log\left(\frac{1}{\delta} \frac{m}{\log(m/\delta)}\right) \\ \frac{\log(4/\delta)}{n} \leq \frac{\log(m/\delta)}{m} \\ 12 K^2 \frac{\log(m/\delta)}{m} \leq \|C\|_{\mathcal{L}(\mathcal{H})} \end{cases}.$$

As $m \leq n$, the third condition is satisfied as soon as $m \geq 4$. Moreover, with this choice of $m$, we have $\log(m/\delta) > 1$, hence the second constraint always holds and we are left with

$$m \geq \max(67, 12K^2 \|C\|_{\mathcal{L}(\mathcal{H})}^{-1}) \log\left(\frac{m}{\delta}\right)$$

Note that this can be rewritten $y - c \log(y) \geq 0$ with $y = m/\delta$ and $c = \max(67, 12K^2 \|C\|_{\mathcal{L}(\mathcal{H})}^{-1})/\delta$.

By looking at asymptotic rates, it is clear that it is always possible to satisfy this equation by taking $m$ large enough. To find the threshold, we look for real solutions of $\frac{-y}{c} + \log\left(\frac{-y}{c}\right) + \log(-c) = 0$. Taking the exponential we get $\exp\left(\frac{-y}{c}\right)\left(\frac{-y}{c}\right) < \left(-\frac{1}{c}\right)$. Since $\delta < \max(67, \|C\|_{\mathcal{L}(\mathcal{H})}^{-1})/e$ always holds, the previous inequality is satisfied when $m \in -W(-\frac{1}{c})\delta c$, where $W$ denote the (multivariate) Lambert's $W$ function. This concludes the proof. $\square$

*Proof of Corollary 4.1:* Under the conditions of Theorem 4.1, the latter gives

$$\|\mu - \hat{\mu}_m\| \leq \frac{c_1}{\sqrt{n}} + \frac{c_2}{m} + \frac{c_3 \sqrt{\log(m/\delta)}}{m} \sqrt{\mathcal{N}\left(\frac{12K^2 \log(m/\delta)}{m}\right)}$$

We split the proof of the corollary in two parts, one for each family of assumptions on the effective dimension $\mathcal{N}$ (polynomial and logarithmic growth).

**Polynomial growth assumption:** $\mathcal{N}(\lambda) \leq c_\gamma \lambda^{-\gamma}$. Setting $m = n^{1/(2-\gamma)} \log(n/\delta)$, we get

$$\|\mu - \hat{\mu}_m\| \leq \frac{c_1}{\sqrt{n}} + \frac{c_2}{m} + c_3 (12K^2)^{-\gamma/2} \frac{\log(m/\delta)^{\frac{1-\gamma}{2}}}{m^{\frac{2-\gamma}{2}}} \tag{8}$$

$$\leq \frac{1}{\sqrt{n}} \left(c_1 + \frac{c_2 n^{1/2}}{\log(n/\delta) n^{1/(2-\gamma)}} + \frac{c_3 (12K^2)^{-\gamma/2}}{\log(n/\delta)^{1/2}}\right) = O\left(\frac{1}{\sqrt{n}}\right) \tag{9}$$

**Logarithmic growth assumption:** $\mathcal{N}(\lambda) \leq \log(1 + c/\lambda)/\beta$. Since $m \geq \frac{12K^2 \log(m/\delta)}{c}$, using the fact that $\log(1+x) \leq \log(2x)$ for $x > 1$ we get

$$\frac{\sqrt{\log(m/\delta)}}{\sqrt{\beta}m} \sqrt{\log\left(1 + \frac{cm}{12K^2 \log(m/\delta)}\right)} \leq \frac{1}{\sqrt{\beta}m} \sqrt{\log(m/\delta) \log\left(\frac{cm}{6K^2 \log(m/\delta)}\right)}$$

$$\leq \frac{1}{\sqrt{\beta}m} \log(m \max(1/\delta, c/(6K^2)))$$

One can easily check that the choice $m = \sqrt{n} \log(\sqrt{n} \max(1/\delta, c/(6K^2)))$ yields the bound

$$\frac{\sqrt{\log(m/\delta)}}{\sqrt{\beta}m} \sqrt{\log\left(1 + \frac{cm}{12K^2 \log(m/\delta)}\right)} \leq \frac{2}{\sqrt{\beta n}}.$$

Plugging the latter bound in (8), we obtain

$$\|\mu - \hat{\mu}_m\| \leq \frac{c_1}{\sqrt{n}} + \frac{c_2}{\sqrt{n} \log(\sqrt{n} \max(1/\delta, c/(6K^2)))} + \frac{2c_3}{\sqrt{\beta n}} = O\left(\frac{1}{\sqrt{n}}\right)$$

This concludes the proof.

$\square$

## D. Maximum Mean Discrepancy

**Theorem D.1 (MMD).** *Let Assumption 4.1 hold. Furthermore, assume that for $k \in \{1, 2\}$, the data points $X_1^k, \ldots, X_{n_k}^k$ are drawn i.i.d. from the distribution $\rho_k$ and that $m_k \leq n_k$ sub-samples $\tilde{X}_1^k, \ldots, \tilde{X}_{m_k}^k$ are drawn uniformly with replacement from the dataset $\{X_1^k \ldots, X_{n_k}^k\}$. Then, for any $\lambda_k \in ]0, \|C_k\|_{\mathcal{L}(\mathcal{H})}]$ and $\delta \in ]0, 1[$, with probability at least $1 - 2\delta$*

$$|\mathrm{MMD}(\rho_1, \rho_2) - \widehat{\mathrm{MMD}}| \leq \sum_{k \in \{1,2\}} \frac{2K\sqrt{2\log(6/\delta)}}{\sqrt{n_k}} + \sqrt{\lambda_k}\left(\frac{4\sqrt{3\mathcal{N}_\infty^{(\rho_k)}(\lambda_k)}\log(12/\delta)}{m_k} + 6\sqrt{\frac{\mathcal{N}^{(\rho_k)}(\lambda)\log(12/\delta)}{m_k}}\right),$$

*provided that, for $k \in \{1, 2\}$,*

- $m_k \geq \max(67, 5\mathcal{N}_\infty^{(\rho_k)}(\lambda_k)) \log \frac{12K^2}{\lambda_k \delta}$,

- $\lambda_k n_k \geq 12K^2 \log(4/\delta)$.

*Proof of Theorem D.1:* By definition of $\widehat{\mathrm{MMD}}$, a reverse triangle inequality followed by a triangle inequality yields

$$\begin{aligned}
|\mathrm{MMD}(\rho_1, \rho_2) - \widehat{\mathrm{MMD}}| &= \left| \|\mu(\rho_1) - \mu(\rho_2)\| - \|P_1\hat{\mu}(\rho_1) - P_1\hat{\mu}(\rho_2)\| \right| \\
&\leq \|\mu(\rho_1) - \mu(\rho_2) - P_1\hat{\mu}(\rho_1) + P_2\hat{\mu}(\rho_2)\| \\
&\leq \sum_{k \in \{1,2\}} \|\mu(\rho_k) - P_k\hat{\mu}(\rho_k)\|.
\end{aligned}$$

We now apply twice Lemma C.1, and conclude via a union bound. $\square$

## E. Concentration inequalities

This section contains concentration results that we rely on to prove our main result.

The first lemma provides a high-probability control on the norm of the average of bounded random variables taking values in a separable Hilbert space.

**Lemma E.1.** *Let $X_1, \ldots, X_n$ be i.i.d. random variables on a separable Hilbert space $(\mathcal{X}, \|\cdot\|)$ such that $\sup_{i=1,\ldots,n} \|X_i\| \leq A$ almost surely, for some real number $A > 0$. Then, for any $\delta \in (0, 1)$, it holds with probability at least $1 - \delta$ that*

$$\left\| \frac{1}{n} \sum_{i=1}^n X_i \right\| \leq A \frac{\sqrt{2\log(2/\delta)}}{\sqrt{n}}.$$

The proof of Lemma E.1 relies on (Pinelis 1994, Theorem 3.5) which we recall now for clarity of exposition.

**Lemma E.2.** *Let $M = (M_i)_{i \in \mathbb{N}}$ be a martingale on a $(2, D)$-smooth separable Banach space $(\mathcal{X}, \|\cdot\|)$. Define $\sum_{j=1}^\infty \mathrm{ess}\sup\|M_j - M_{j-1}\|^2 \leq b_*^2$, for some real number $b_* > 0$. Then, for all $r \geq 0$,*

$$\Pr\left[\sup_{j \in \mathbb{N}} \|M_j\| \geq r\right] \leq 2\exp\left(-\frac{r^2}{2D^2 b_*^2}\right).$$

We now prove Lemma E.1.

*Proof of Lemma E.1:* Since $\mathcal{X}$ is a Hilbert space, it is 2-smooth with 2-smoothness constant $D = 1$. We define the martingale $(M_n)_{n \in \mathbb{N}}$ as $M_0 = 0$, $M_k = \sum_{1 \leq i \leq k} X_k$ for $1 \leq k \leq n$ and $M_k = M_n$ for $k \geq n$, so that

$$d_k := M_k - M_{k-1} = \begin{cases} X_k, & \text{if } 1 \leq k \leq n \\ 0, & \text{otherwise} \end{cases}.$$

As a consequence, we have $\sum_{j=1}^{\infty} \operatorname{ess\,sup} \|d_j\|^2 = \sum_{j=1}^{n} \operatorname{ess\,sup} \|X_j\|^2 \leq nA^2$. Applying Pinelis' inequality (Lemma E.2) with $b_*^2 = nA^2$ yields

$$\Pr\left[\left\|\frac{1}{n}\sum_{i=1}^{n} X_i\right\| > \epsilon\right] = \Pr\left[\|M_n\| > n\epsilon\right] \leq \Pr\left[\sup_{1 \leq j \leq n} \|M_j\| > n\epsilon\right] \leq 2\exp\left(-\frac{n\epsilon^2}{2A^2}\right).$$

We get the desired result by choosing $\epsilon = A\sqrt{2\log(2/\delta)}n^{-1/2}$. $\qquad\square$

The next result is a Bernstein-type inequality for random vectors defined in a Hilbert space.

**Lemma E.3 (Bernstein inequality for Hilbert space-valued random vectors).** *Let $X_1, \ldots, X_n$ be i.i.d. random variables in a Hilbert space $(\mathcal{H}, \|\cdot\|)$ such that*

- $\forall i \in [n], \mathbf{E}X_i = \mu,$
- $\exists \sigma > 0, \exists H > 0, \forall i \in [n], \forall p \geq 2, \ \mathbf{E}\|X_i - \mu\|^p \leq \frac{1}{2}p!\sigma^2 H^{p-2}.$

*Then, for any $\delta \in ]0, 1[$, we have with probability at least $1 - \delta$,*

$$\left\|\frac{1}{n}\sum_{i=1}^{n} X_i - \mu\right\| \leq \frac{2H\log(2/\delta)}{n} + \sqrt{\frac{2\sigma^2\log(2/\delta)}{n}}.$$

*Proof of Lemma E.3:* Fix a confidence level $\delta \in (0, 1)$. Applying (Yurinsky 1995, Theorem 3.3.4) on the i.i.d. centered random variables $\xi_i = X_i - \mu$ with $B^2 = \sigma^2 n$, we get

$$\Pr\left[\left\|\frac{1}{n}\sum_{j=1}^{n} \xi_j - \mu\right\| \geq t\right] \leq \Pr\left[\max_{1 \leq k \leq n} k\left\|\frac{1}{k}\sum_{j=1}^{k} \xi_j - \mu\right\| \geq \left(\frac{tn}{B}\right)B\right] \leq 2\exp\left(-\frac{1}{2}\frac{(tn)^2}{B^2}\left(1 + \frac{tHn}{B^2}\right)^{-1}\right).$$

The RHS of the above is smaller than $\delta$ if and only if

$$t^2 n^2 - t(2Hn\log(2/\delta)) - 2B^2\log(2/\delta) \geq 0.$$

Denoting $\Delta = 4H^2 n^2 \log(2/\delta)^2 + 8n^2 B^2 \log(2/\delta) > 0$, this holds in particular if $t \geq \frac{H\log(2/\delta)}{n} + \frac{\sqrt{\Delta}}{2n^2}$, and thus a fortiori (using $\sqrt{\Delta} \leq \sqrt{4H^2 n^2 \log(2/\delta)^2} + \sqrt{8n^2 B^2 \log(2/\delta)}$) when

$$t \geq \frac{2H\log(2/\delta)}{n} + \sqrt{\frac{2\sigma^2\log(2/\delta)}{n}}.$$

$\qquad\square$

The following lemma provides a Bernstein-type bound for the empirical mean of Hilbert space-valued centered random variables 'whitened" by regularized linear operator.

**Lemma E.4.** *Let $X_1, \ldots, X_n$ be i.i.d. random variables taking values in a separable Hilbert space $(\mathcal{H}, \langle \cdot, \cdot \rangle)$ with associated norm $\|\cdot\|$. We denote their mean by $\mu_X := \mathbf{E}X_1$ and their covariance by $C := \mathbf{E}[X_1 \otimes X_1]$.*

*Let $Q : \mathcal{H} \to \mathcal{H}$ be a linear operator. For any $\lambda > 0$ and $\delta \in ]0,1[$, it holds with probability at least $1 - \delta$ that*

$$\left\| Q_\lambda^{-1/2} \left( \frac{1}{n} \sum_{i=1}^n X_i - \mu_X \right) \right\| \leq \frac{4 \operatorname{ess\,sup} \left\| Q_\lambda^{-1/2} X_1 \right\| \log(2/\delta)}{n} + \sqrt{\frac{4 \operatorname{tr}(Q_\lambda^{-1} C) \log(2/\delta)}{n}}.$$

*Proof of Lemma E.4:*   To prove the stated result we will apply Lemma E.3 on the random variables $(\zeta_i)_{1 \leq i \leq n}$ defined by $\zeta_i = Q_\lambda^{-1/2} X_i$. Let $N_Q(\lambda) = \operatorname{tr}(Q_\lambda^{-1} C)$ and $N_{Q,\infty}(\lambda) := \operatorname{ess\,sup} \left\| Q_\lambda^{-1/2} X_1 \right\|$.

For any index $1 \leq i \leq n$, we have $\mathbf{E}\zeta_1 = Q_\lambda^{-1/2} \mu_X$,

$$\operatorname{ess\,sup} \| \zeta_i - \mathbf{E}[\zeta_i] \| \leq 2 \operatorname{ess\,sup} \| \zeta_i \| = 2 N_\infty(\lambda)^{1/2},$$

and,

$$\begin{aligned}
\mathbf{E}\| \zeta_i - \mathbf{E}[\zeta_i] \|^2 &= \operatorname{tr}(\mathbf{E}\langle \zeta_i - \mathbf{E}[\zeta_i], \zeta_i - \mathbf{E}[\zeta_i] \rangle) \\
&= \operatorname{tr}(\mathbf{E}[(\zeta_i - \mathbf{E}[\zeta_i]) \otimes (\zeta_i - \mathbf{E}[\zeta_i])]) \\
&= \operatorname{tr}(\mathbf{E}[\zeta_i \otimes \zeta_i] - \mathbf{E}\zeta_i \otimes \mathbf{E}\zeta_i) \\
&\leq \operatorname{tr}(\mathbf{E}[\zeta_i \otimes \zeta_i]) \\
&= \operatorname{tr}(Q_\lambda^{-1/2} C Q_\lambda^{-1/2}) \\
&= N_Q(\lambda).
\end{aligned}$$

Moreover, for any $p \geq 2$,

$$\begin{aligned}
\| \zeta_i - \mathbf{E}[\zeta_i] \|^p &\leq (\mathbf{E}\| \zeta_i - \mathbf{E}[\zeta_i] \|^2)(\operatorname{ess\,sup} \| \zeta_i - \mathbf{E}[\zeta_i] \|^{p-2}) \\
&\leq \tfrac{1}{2}(2 N_Q(\lambda))(2 N_{Q,\infty}(\lambda)^{1/2})^{p-2} \\
&\leq \tfrac{1}{2} p! (2 N_Q(\lambda))(2 N_{Q,\infty}(\lambda)^{1/2})^{p-2}.
\end{aligned}$$

The result follows from Lemma E.3 with constants $\sigma^2 = 2 N_Q(\lambda)$ and $H = 2 N_{Q,\infty}(\lambda)^{1/2}$.   □

Lemma E.5 is a specialization of Lemma E.4 to bound the last term appearing in Lemma 4.1 in our setting of Nyström uniform sampling.

**Lemma E.5.** *Assume that the $m \geq 1$ Nyström landmarks are sampled uniformly with replacement from the dataset $X_1, \ldots, X_n$. If $0 < \lambda \leq \|C\|_{\mathcal{L}(\mathcal{H})}$ and $\lambda n \geq 12 K^2 \log(4/\delta)$, it holds with probability at least $1 - \delta$,*

$$\left\| C_\lambda^{-1/2}(\hat{\mu} - \tilde{\mu}_m) \right\| \leq \frac{4 \sqrt{\mathcal{N}_\infty(\lambda)} \log(4/\delta)}{m} + \sqrt{\frac{12 \mathcal{N}(\lambda) \log(4/\delta)}{m}}.$$

*Proof of Lemma E.5:*   Fix the desired confidence level $\delta \in (0,1)$. Let us beginning by conditioning w.r.t. to the dataset $X_1, \ldots, X_n$. As the landmarks are assumed to be drawn i.i.d., we can apply Lemma E.4 with $Q = C$ on the i.i.d. random variables $h_j := \phi(\tilde{X}_j), 1 \leq j \leq m$, satisfying $\mathbf{E}[h_1] = \hat{\mu}$, $\mathbf{E}[h_1 \otimes h_1] = \hat{C}_n$ and $\operatorname{ess\,sup} \| C_\lambda^{-1/2} h_1 \|^2 \leq \mathcal{N}_\infty(\lambda)$: it holds with probability at least $1 - \delta/2$ (over the drawing of the landmarks) that

$$\left\| Q_\lambda^{-1/2}(\mu_X - \hat{\mu}_X) \right\| \leq \frac{4 \sqrt{\mathcal{N}_\infty(\lambda)} \log(4/\delta)}{m} + \sqrt{\frac{4 \operatorname{tr}(C_\lambda^{-1} \hat{C}_n) \log(4/\delta)}{m}}.$$

Then, since we assumed $\lambda \leq \|C\|_{\mathcal{L}(\mathcal{H})}$ and $\lambda n \geq 12 K^2 \log(4/\delta)$, Lemma E.6 ensures that $\operatorname{tr}(C_\lambda^{-1} \hat{C}_n) \leq 3 \mathcal{N}(\lambda)$ with probability at least $1 - \delta/2$ w.r.t. the dataset $X_1, \ldots, X_n$.

Finally, since the drawing of dataset and that of the indexes of the landmark are independent, the claimed bound holds with probability at least $(1 - \delta/2)(1 - \delta/2) \geq 1 - \delta$.   □

The next lemma

**Lemma E.6.** *Let $\delta > 0$, $\lambda > 0$ and $n \in \mathbb{N}$ be such that $\lambda \leq \|C\|_{\mathcal{L}(\mathcal{H})}$ and $n \geq 12\mathcal{N}_\infty(\lambda)\log(2/\delta)$. Then it holds with probability at least $1 - \delta$ that*

$$\mathrm{tr}(C_\lambda^{-1}\hat{C}_n) \leq 3\mathcal{N}(\lambda).$$

*Proof of Lemma E.6:* Let us control the deviation of $\mathrm{tr}(C_\lambda^{-1}\hat{C}_n)$ from its expectation $\mathcal{N}(\lambda)$. We have

$$\mathrm{tr}(C_\lambda^{-1}\hat{C}_n) - \mathcal{N}(\lambda) = \mathrm{tr}(C_\lambda^{-1}(\hat{C}_n - C)) = \frac{1}{n}\sum_{i=1}^n \xi_i - \mathbf{E}[\xi_i],$$

where we define $\xi_i := \mathrm{tr}(C_\lambda^{-1}\phi(X_i) \otimes \phi(X_i)), i = 1, \ldots, n$. The random variables $\xi_i, 1 \leq i \leq n$, satisfy

$$|\xi_i - \mathbf{E}[\xi_i]| = \left|\mathrm{tr}(C_\lambda^{-1}(\phi(X_i) \otimes \phi(X_i) - C))\right| \leq \left\|C_\lambda^{-1/2}\phi(X_i)\right\|^2 + \mathcal{N}(\lambda) \leq 2\mathcal{N}_\infty(\lambda)$$

and

$$\mathbf{E}[(\xi_i - \mathbf{E}[\xi_i])^2] = \mathbf{E}[\xi_i^2] - (\mathbf{E}\xi_i)^2 \leq \mathrm{ess\,sup}\,|\xi_i|\,\mathbf{E}[\xi_i] \leq 2\mathcal{N}_\infty(\lambda)\mathcal{N}(\lambda).$$

Lemma E.3 with $H = 2\mathcal{N}_\infty(\lambda)$ and $\sigma^2 = 2\mathcal{N}_\infty(\lambda)\mathcal{N}(\lambda)$ ensures that with probability at least $1 - \delta$,

$$|\mathrm{tr}(C_\lambda^{-1}\hat{C}_n) - \mathcal{N}(\lambda)| \leq \frac{4\mathcal{N}_\infty(\lambda)\log(2/\delta)}{n} + \sqrt{\frac{4\mathcal{N}_\infty(\lambda)\mathcal{N}(\lambda)\log(2/\delta)}{n}}.$$

Since $\lambda \leq \|C\|_{\mathcal{L}(\mathcal{H})}$, we have $\mathcal{N}(\lambda) = \mathrm{tr}(CC_\lambda^{-1}) \geq \|CC_\lambda^{-1}\|_{\mathcal{L}(\mathcal{H})} = \frac{\|C\|_{\mathcal{L}(\mathcal{H})}}{\|C\|_{\mathcal{L}(\mathcal{H})} + \lambda} \geq 1/2$. Furthermore, using the assumption $n \geq 12\mathcal{N}_\infty(\lambda)\log(2/\delta)$, it holds with probability at least $1 - \delta$,

$$\mathrm{tr}(C_\lambda^{-1}\hat{C}_n) \leq \mathcal{N}(\lambda)\left(1 + \frac{1}{3\mathcal{N}(\lambda)} + \sqrt{\frac{1}{3\mathcal{N}(\lambda)}}\right) \leq \mathcal{N}(\lambda)\left(1 + \frac{2}{3} + \sqrt{\frac{2}{3}}\right) \leq 2.5\mathcal{N}(\lambda).$$

$\square$

# F. Nyström approximation result

To control the term involving $P_m^\perp$, we rely on the following lemma from Rudi et. al (Rudi et al. 2015, Lemma 6).

**Lemma F.1 (Uniform Nyström approximation).** *When the set of $m$ landmarks is drawn uniformly from all partitions of cardinality $m$, for any $\lambda \in ]0, \|C\|_{\mathcal{L}(\mathcal{H})}]$ we have*

$$\|P_m^\perp(C + \lambda I)^{1/2}\|_{\mathcal{L}(\mathcal{H})}^2 \leq 3\lambda$$

*with probability at least $1 - \delta$ provided*

$$m \geq \max(67, 5\mathcal{N}_\infty(\lambda))\log\frac{4K^2}{\lambda\delta}.$$

# G. Experiments

We provide here two expressions which are used to compute the exact error in Section 6.1.

We denote by $\mathcal{N}_x(\mu, \Gamma)$ the evaluation of the Gaussian density of mean $\mu$ and covariance matrix $\Gamma$ at point $x$.

Let $d \geq 1$, $\mu, \tilde{\mu} \in \mathbb{R}^d$, and $\Gamma = diag(\gamma_1, \ldots, \gamma_d), \tilde{\Gamma} = diag(\tilde{\gamma}_1, \ldots, \tilde{\gamma}_d)$. For any $x, \tilde{x} \in \mathbb{R}^d$, using (Petersen et al.

2008, Eq. 371),

$$\mathbb{E}_{x \sim \mathcal{N}(\mu, \Gamma)} k(x, \tilde{x}) = \int_{\mathbb{R}^d} (2\pi\tau^2)^{d/2} \mathcal{N}_x(\tilde{x}, \tau^2 I_d) \mathcal{N}_x(\mu, \Gamma) dx$$

$$= (2\pi\tau^2)^{d/2} \mathcal{N}_{\tilde{x}}(\mu, \Gamma + \tau^2 I_d) \int \mathcal{N}_x(m_c, \Sigma_c) d_x$$

$$= (2\pi\tau^2)^{d/2} \mathcal{N}_{\tilde{x}}(\mu, \Gamma + \tau^2 I_d),$$

and

$$\mathbb{E}_{\tilde{x} \sim \mathcal{N}(\tilde{\mu}, \tilde{\Gamma})} \mathbb{E}_{x \sim \mathcal{N}(\mu, \Gamma)} k(x, \tilde{x}) = (2\pi\tau^2)^{d/2} \int_{\mathbb{R}^d} \mathcal{N}_{\tilde{x}}(\mu, \Gamma + \tau^2 I_d) \mathcal{N}_{\tilde{x}}(\tilde{\mu}, \tilde{\Gamma}) dx$$

$$= (2\pi\tau^2)^{d/2} \mathcal{N}_{\tilde{\mu}}(\mu, \Gamma + \tilde{\Gamma} + \tau^2 I_d).$$